\documentclass{article}

\usepackage[preprint]{neurips_2021}

\usepackage[utf8]{inputenc} 
\usepackage[T1]{fontenc}    
\usepackage{hyperref}       
\usepackage{url}            
\usepackage{booktabs}       
\usepackage{amsfonts}       
\usepackage{nicefrac}       
\usepackage{microtype}      
\usepackage{xcolor}         

\usepackage[linesnumbered, ruled]{algorithm2e}
\usepackage{epsfig} 

\usepackage{graphicx}

\usepackage{algpseudocode}
\usepackage{amsmath}
\usepackage{subfig}

\usepackage{amsfonts,amssymb}
\usepackage{multirow}

\usepackage[linesnumbered, ruled]{algorithm2e}

\title{TiKick: Towards Playing Multi-agent Football Full Games from Single-agent Demonstrations}

\author{%
  Shiyu Huang\thanks{Equal contribution} \\
  Tsinghua University\\
  Beijing, China \\
  \texttt{hsy17@mails.tsinghua.edu.cn} \\
   \And
   Wenze Chen\footnotemark[1] \\
   Tsinghua University \\
   Beijing, China \\
   \texttt{cwz19@mails.tsinghua.edu.cn} \\
    \And
   Longfei Zhang \\
   National University of Defense Technology \\
   Changsha, China \\
   \texttt{zhanglongfei@nudt.edu.cn} \\
   \And
   Shizhen Xu \\
   RealAI \\
   Beijing, China \\
   \texttt{shizhen.xu@realai.ai} \\
   \And
   Ziyang Li \\
   Tencent AI Lab \\
   Shenzhen, China \\
   \texttt{tzeyangli@tencent.com} \\
   \And
   Fengming Zhu \\
   Tencent AI Lab \\
   Shenzhen, China \\
   \texttt{fridazhu@tencent.com} \\
   \And
   Deheng Ye \\
   Tencent AI Lab \\
   Shenzhen, China \\
   \texttt{dericye@tencent.com} \\
   \And
   Ting Chen \\
   Tsinghua University \\
   Beijing, China \\
   \texttt{tingchen@tsinghua.edu.cn} \\
   \And
   Jun Zhu \\
   Tsinghua University \\
   Beijing, China \\
   \texttt{dcszj@tsinghua.edu.cn} \\
}

\begin{document}

\maketitle

\begin{abstract}
Deep reinforcement learning (DRL) has achieved super-human performance on complex video games (e.g., StarCraft II and Dota II). 
However, current DRL systems still suffer from challenges of multi-agent coordination, sparse rewards, stochastic environments, etc. 
In seeking to address these challenges, we employ a football video game, e.g., Google Research Football (GRF), as our testbed and develop an end-to-end learning-based AI system (denoted as TiKick\footnote{Codes can be found at \url{https://github.com/TARTRL/TiKick}.}\footnote{Videos available at \url{https://sites.google.com/view/tikick}.}) to complete this challenging task. In this work, we first generated a large replay dataset from the self-playing of single-agent experts, which are obtained from league training. 
We then developed a distributed learning system and new offline algorithms to learn a powerful multi-agent AI from the fixed single-agent dataset. 
To the best of our knowledge, TiKick is the first learning-based AI system that can take over the multi-agent Google Research Football full game, while previous work could either control a single agent or experiment on toy academic scenarios. Extensive experiments further show that our pre-trained model can accelerate the training process of the modern multi-agent algorithm and our method achieves state-of-the-art performances on various academic scenarios.
\end{abstract}

\section{Introduction}

Deep reinforcement learning (DRL) has shown great success in many video games, including the Atari games~\citep{mnih2013playing}, StarCraft II~\citep{vinyals2019grandmaster}, Dota II~\citep{berner2019dota}, etc. However, current DRL systems still suffer from challenges of multi-agent coordination~\citep{rashid2018qmix,mahajan2019maven,yu2021surprising}, sparse rewards~\citep{taiga2019bonus,zhang2020bebold}, stochastic environments~\citep{googlefootball,team2021open}, etc. In seeking to address these challenges, we employ a football video game, e.g., Google Research Football (GRF)~\citep{googlefootball}, as our testbed. 

Even though a lot of work has been done recently~\citep{googlefootball,li2021celebrating,liu2021unifying}, there remain many problems in building agents for the GRF: 
(1) {\bf Multiple Players}:  In the GRF, there are both cooperative and competitive players. For cooperative players, the joint action space is very huge, thus it is hard to build a single agent to control all the players. Moreover, competitive players mean that the opponents are not fixed, thus the agents should be adapted to various opponents. (2) {\bf Sparse Rewards}: The goal of the football game is to maximize the goal score, which can only be obtained after a long time of the perfect decision process. And it is almost impossible to receive a positive reward when starting from random agents. (3) {\bf Stochastic Environments}: The GRF introduces stochasticity into the environment, which means the outcome of taking specific actions is not deterministic. This can improve the robustness of the trained agents but also increases the training difficulties.

To address the aforementioned issues, we develop an end-to-end learning-based AI system (denoted as TiKick) to complete this challenging task. In this work, we first generated a large replay dataset from the self-playing of single-agent experts, which are obtained from league training. 
We then developed a distributed learning system and new offline algorithms to learn a powerful multi-agent AI from the fixed single-agent dataset. 
To the best of our knowledge, TiKick is the first learning-based AI system that can take over the multi-agent Google Research Football full game, while previous work could either control a single agent or experiment on toy academic scenarios. Extensive experiments further show that our pre-trained model can accelerate the training process of the modern multi-agent algorithm and our method can achieve state-of-the-art performances on various academic scenarios.

\section{Background}
In this section, we will first introduce basic notations in multi-agent reinforcement learning (MARL) and then briefly review imitation learning and offline reinforcement learning. We also present the efforts made in the football game AI research field.

\subsection{Multi-Agent Reinforcement Learning}
The multi-agent reinforcement learning can be formalized as Dec-POMDPs~\citep{oliehoek2016concise}. Formally, a Dec-POMDP is represented as a tuple $(\mathcal{A},S,U,T,r,O,G,\gamma)$, where $S$, $U$, $O$ and $\gamma$ are state space, action space, observation space and discount factor, respectively. 
$\mathcal{A}\equiv\{1,...,n\}$ is the set of $n$ agents.
At each time step $t$, each agent $i\in \mathcal{A}$ chooses an action $u^i\in U$ to form a joint action $\mathbf{u}\in \mathbf{U}\equiv U^n$. 
The immediate reward function $r(s,\mathbf{u})$ is the received reward when taking action $\mathbf{u}$ in state $s$ and it is shared by all the agents.
$T(s,\mathbf{u},s'):S\times\mathbf{U}\times S\rightarrow [0,1]$ is the state-transition function, which defines the probability of the succeeding state $s'$ after taking action $\mathbf{u}$ in state $s$. In Dec-POMDP, each agent can only have access to the partially observable observations $o\in O$ according to the observation function $G(s,i): S\times\mathcal{A}\rightarrow O$. Each agent uses a policy $\pi^i(u^i|o^i_{1:t})$ to produce its action $u^i$ from its local historical observations $o^i_{1:t}$. $u^{-i}$ is used to denote the action of all the agents other than $i$ and follow a similar convention for the policies $\pi^{-i}$. The goal is to learn a joint policy $\pi$ to 
maximize the discounted accumulated reward $\mathbb{E}_{s_t,\mathbf{u}_t}[\sum_t \gamma^t r(s_t,\mathbf{u}_t)]$. 
%



\begin{figure*}[t]
\subfloat[Multi-agent Full Game]{\begin{centering}
\includegraphics[width=0.31\linewidth]{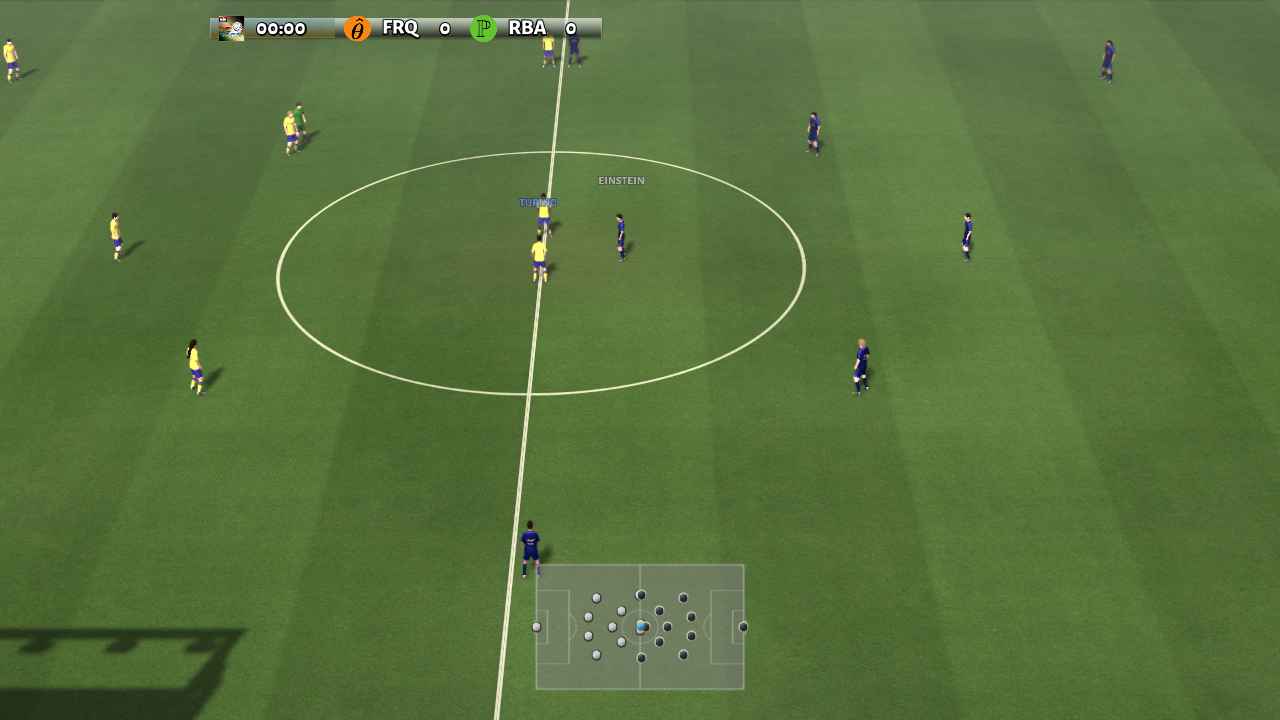}
\end{centering}
}\subfloat[Run to Score with Keeper]{\begin{centering}
\includegraphics[width=0.31\linewidth]{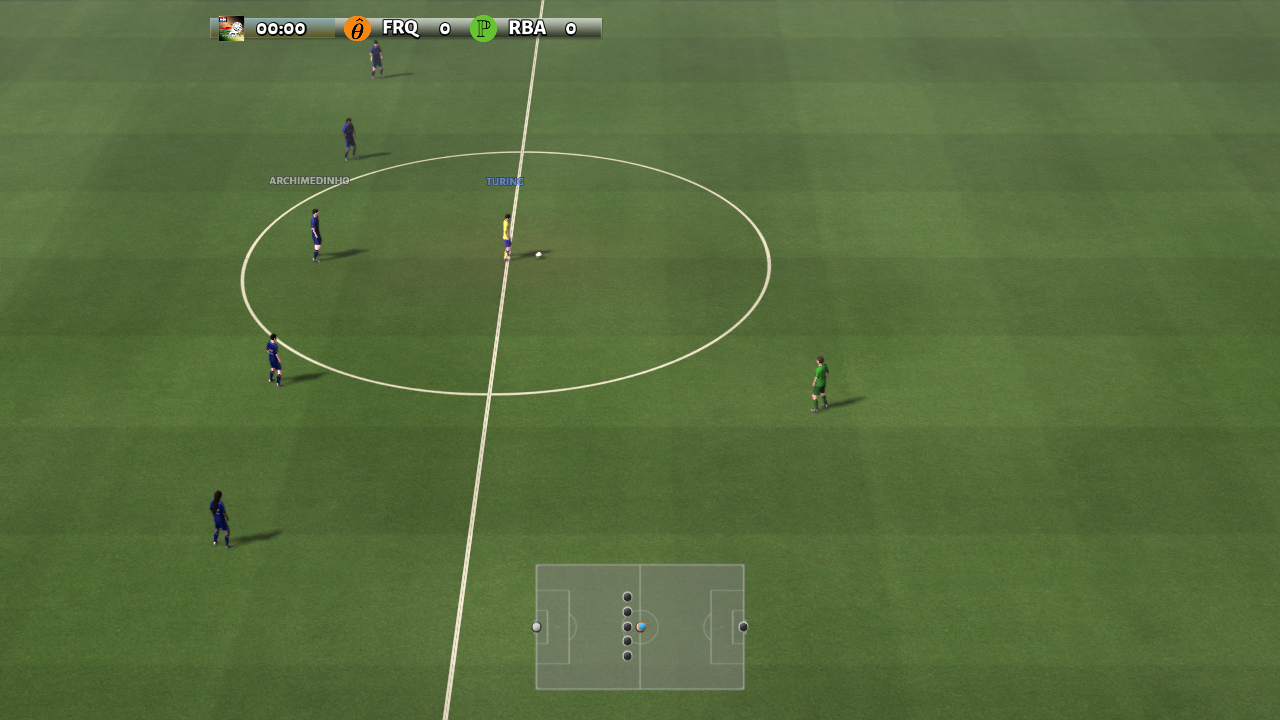}
\end{centering}
}\subfloat[Run, Pass and Shoot with Keeper]{\begin{centering}
\includegraphics[width=0.31\linewidth]{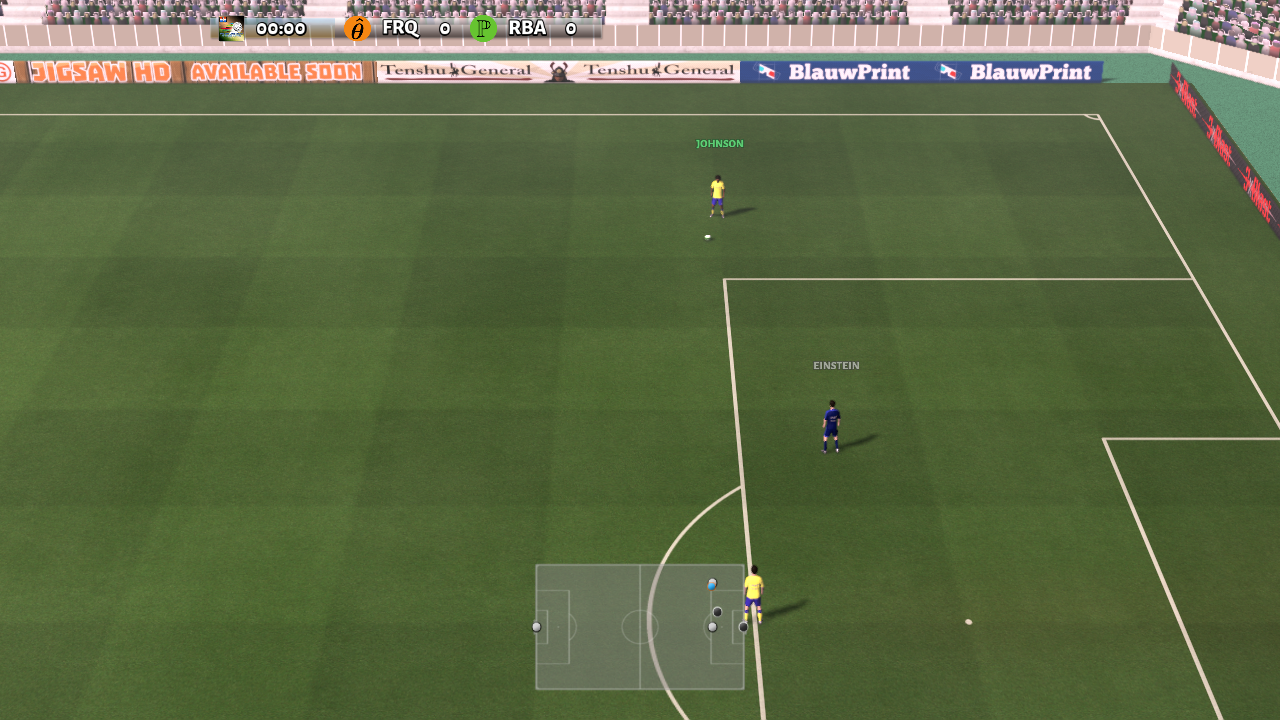}
\end{centering}
}

\subfloat[Corner]{\begin{centering}
\includegraphics[width=0.31\linewidth]{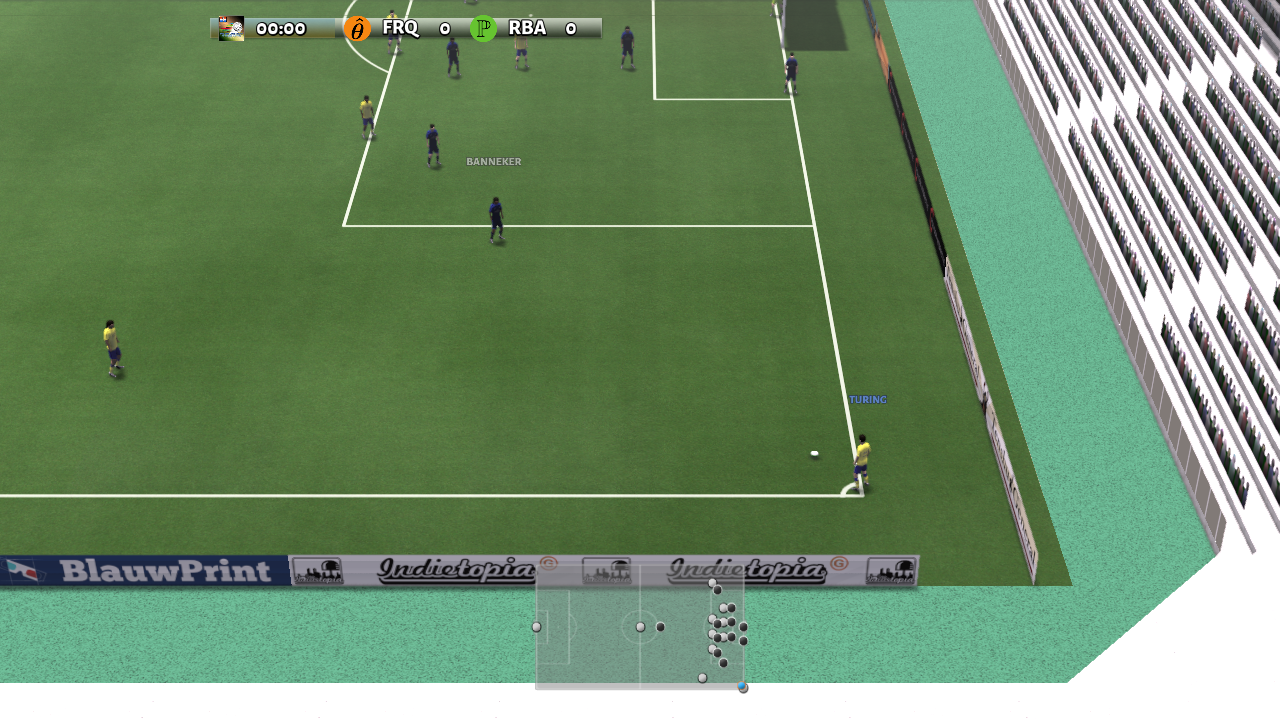}
\end{centering}
}\subfloat[3 vs 1 with Keeper]{\begin{centering}
\includegraphics[width=0.31\linewidth]{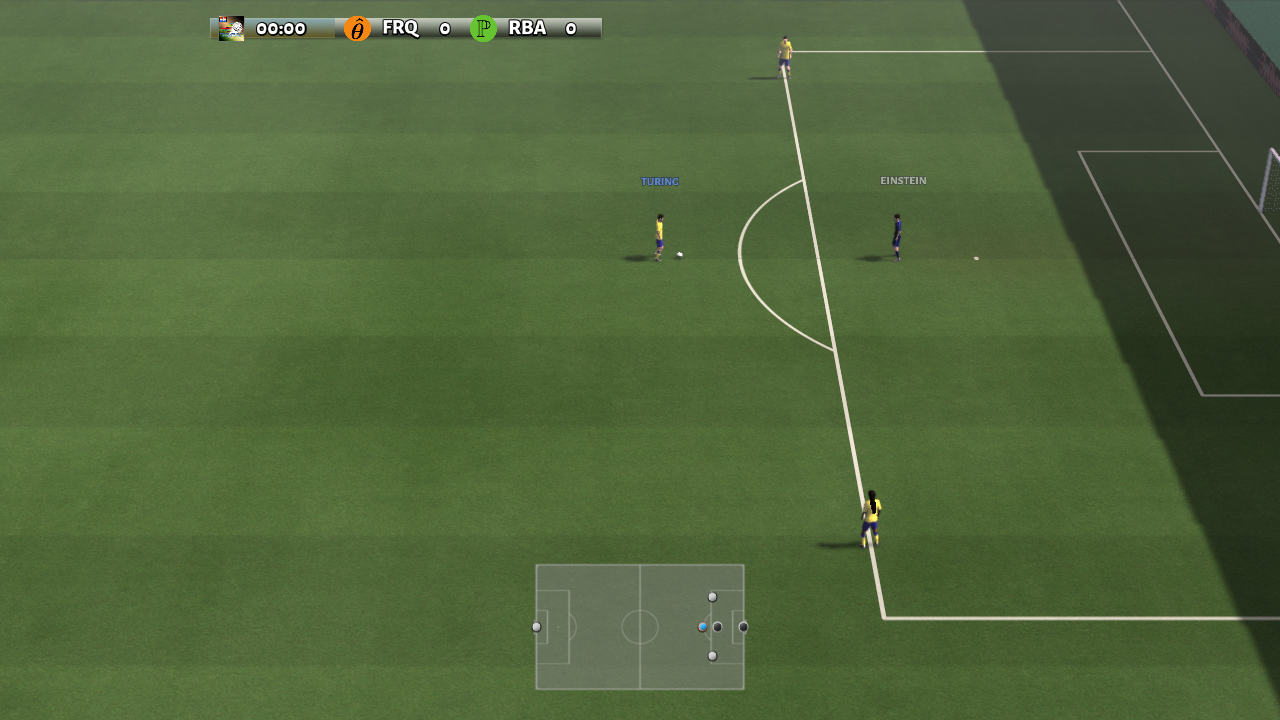}
\end{centering}
}\subfloat[Hard Counter-attack]{\begin{centering}
\includegraphics[width=0.31\linewidth]{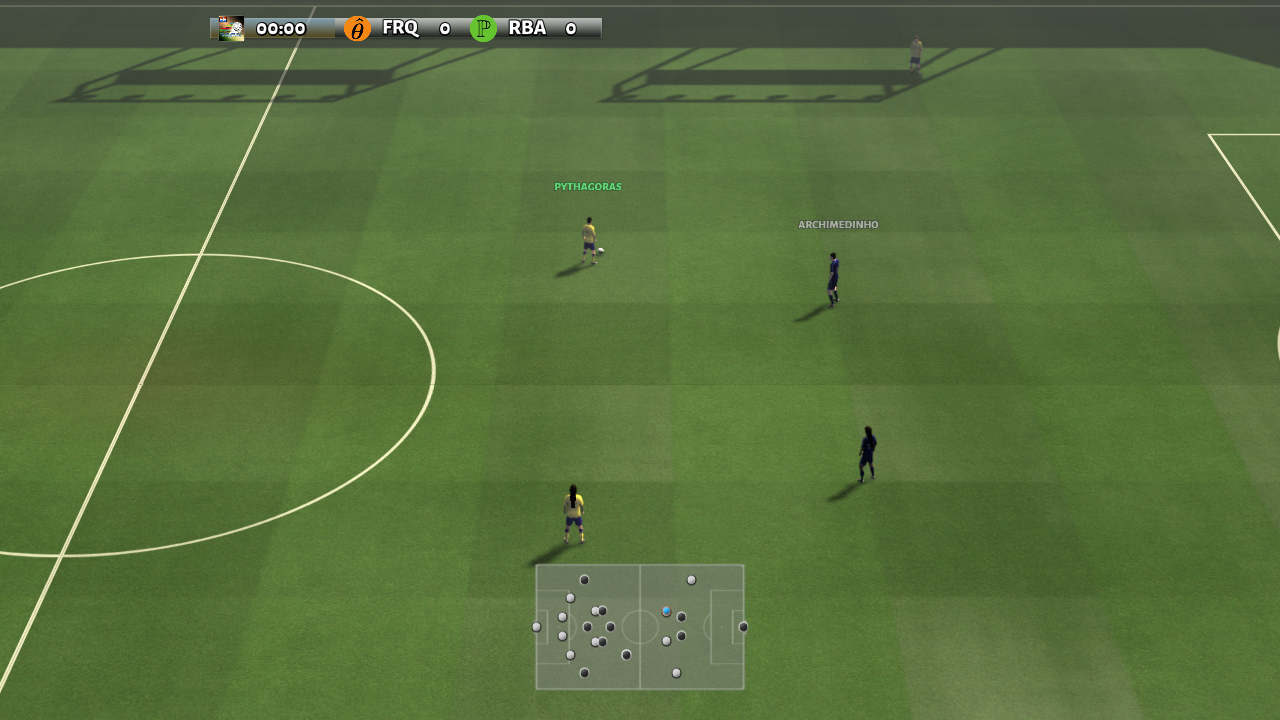}
\end{centering}
}
\caption{Google Research Football screenshots. (a) shows the multi-agent full game, which requires all the ten players to be controlled. (b)-(f) show the five GRF academic scenarios, which require fewer players to be controlled and are easier than the full game.}
\label{fig:screenshots}
\end{figure*}
\subsection{Imitation Learning and Offline Reinforcement Learning}
Training agents from pre-collected dataset has long been studied~\citep{ho2016generative,kang2018policy,oh2018self,peng2019advantage,fu2020d4rl,qin2021neorl}. \cite{peng2019advantage} used standard supervised learning methods to train reinforcement learning agents, which aimed to maximize the expected improvement. Besides, ~\cite{oh2018self,vinyals2019grandmaster} tried to develop new algorithms to learn policies from past good experiences. Moreover, \cite{ho2016generative} proposed a generative adversarial training method (i.e., GAIL) to train a policy to mimic the behavior policy. And then \cite{song2018multi} extended GAIL to multi-agent scenarios. More recently, offline reinforcement learning has attracted attention because it intends to learn a better policy than the behavior policies which generate the dataset~\citep{fu2020d4rl,gulcehre2020rl,qin2021neorl}. Although much work has been proposed, such as model-based methods~\citep{kidambi2020morel,yu2020mopo}, policy constraint methods~\citep{fujimoto2019off,wu2019behavior}, value regularization methods~\citep{kumar2020conservative} and uncertainty-based methods~\citep{agarwal2020optimistic}, there is no work on learning multiple cooperative agents with the incomplete (or partially observable) dataset. And our work tries to build a powerful AI system for the multi-agent football full game from single-agent demonstrations.

\subsection{Football Games and Football Game AIs}
Many researchers have developed agents for other football games (e.g., \emph{RoboCup Soccer Simulator}~\citep{kitano1997robocup} and the \emph{DeepMind MuJoCo Multi-Agent Soccer Environment}~\citep{liu2019emergent,liu2021motor}). Different from Google Research Football, these environments focus more on low-level control of a physics simulation of robots, while GRF focuses on high-level actions. In Google Research Football Competition 2020~\citep{kaggle}, an RL-based agent, named WeKick~\citep{wekick}, took first place. WeKick utilized imitation learning, the multi-head value trick, and distributed league training to achieve the top performance on the GRF. However, it is a single-agent AI that can not be extended to multi-agent control. And \cite{liu2021unifying} also proposed a new algorithm for the GRF with single-agent control. Instead, our work tries to build a powerful game AI for multi-agent control on the GRF. More recently, ~\cite{li2021celebrating} proposed a value-based multi-agent algorithm (i.e., CDS) for GRF with mutual information maximization between agents. However, they only conducted experiments on toy academic scenarios. Compared with CDS, our AI system can achieve much better performances and lower sample complexity on these academic scenarios without the mutual information maximization process, and our method can even play the GRF full game.

\section{Methodology}
In this section, we will introduce the TiKick algorithm in detail, including our observation, model design, and our proposed offline reinforcement learning method.

\subsection{Observation and Model Design}
\label{observaion_design}
Design the observation is the first step for building a DRL model. The Google Research Football simulator originally provide three types of observations. The first type is the pixel-level representation, which consists of a $1280\times 720$ RGB image corresponding to the rendered screen. The first type is the \emph{Super Mini Map} (SMM), which consists of four $72\times 96$ matrices encoding information about the current situation. The third type is the vector representation, which provides a compact encoding and consists of a $115$-dimensional vector. The pixel-level and SMM representations need more burdensome deep models to extract hidden features, which results in lower training speed and higher memory consumption. Thus, we use the vector representation as the input of our deep model. Furthermore, we extend the standard $115$-dimensional vector with auxiliary features, such as relative poses among teammates and opponents, offside flags to mark potential offside teammates, etc. Finally, we obtain a $268$-dimensional vector as the input. More details about the observation can be found in the Appendix.

To extract hidden features from the raw input, we build a deep model with four fully connected layers and GRU cells~\citep{cho2014learning}. All the hidden layers are followed by a ReLU, except for the last output layer and all the hidden sizes are set to $256$. The learning rate is set to $1e-4$ and is fixed during the training. 
Network parameters are initialized with the orthogonal matrix~\citep{saxe2013exact} and Adam optimizer~\citep{kingma2014adam} is used for the parameter update. More details about the network structure and hyper-parameters can be found in the Appendix.

\subsection{Offline RL Training with Single-agent Dataset}
In this part, we will introduce how we collect the single-agent dataset and how to utilize this dataset to learn multi-agent control models.

\subsubsection{Single-agent Dataset Collection}
To collect an expert single-agent dataset, we first obtain a single-agent AI, denoted as WeKick, from self-play league training. WeKick took first place at the Google Research Football Competition 2020~\citep{kaggle} and it is the most powerful football AI in the world until now. We let WeKick play with itself and store all the battle data, including raw observations, actions, and rewards. During the self-playing, only the designated player at one side can be controlled by the WeKick, and the designated player is not fixed and is changed automatically according to the build-in strategy. The game will last for $3,000$ steps for each round (or episode). At last, we collected $21,947$ episodes from the self-play. This dataset will then be used for training our multi-agent offline RL model and other offline RL baselines.

\subsubsection{Multi-agent Offline RL Training}
\label{sec:offline}

The dataset is collected from single-agent control, so it is easy to train a single-agent model with behavior cloning. However, such trained models can not be applied to multi-agent control and many problems will be raised. For example, we find that if we control all the ten players on the court with the same trained model, all the players will be huddled together because all the players tend to get the ball. In single-agent playing, we only need to control the player who is closest to the ball most time. And the designated player is dynamically changed which means the observation inputs for the model are switched between different players. However, for the multi-agent control, we need to control the player who is far from the ball and the observation inputs for each control model are always from a specific player. To handle the differentiation between single-agent and multi-agent playing, we carefully designed the observations, actions, and our learning algorithm. 

As for the observations, we construct a $268$-dimensional vector (as described in Section~\ref{observaion_design}) from the raw observation for each player. As for the actions, we extend the original 19 discrete actions to 20 discrete actions by adding an extra build-in action. All the non-designated players will be assigned the build-in action. When the player takes the build-in action, the player will behave like a build-in agent. Currently, such build-in agents are obtained from a rule-based tactic, and we will try to convert it to a learning-based controller in future work. To be clear, the build-in action is only used for the full game and dropped for all the academic scenarios.

The most direct way to build a multi-agent AI is to train behavior-cloning models with the pre-processed data. The observations-state pair is represented as $\{(o_{1:t}^i,u_t^i)\}_{i=1}^{n}$, where $o_{1:t}^i$ is the observation history of agent $i$, $u_t^i$ is the action of agent $i$ at time-step $t$ and $n$ is the agent number ($n=10$ in the GRF full game). Our goal is to train a parametric neural network policy $\pi_\theta(u^i|o_{1:t}^i)$ to mimic the strategy implied in the data, where $\theta$ is the parameter of the policy network and all the agents share the same parameter. The policy will be updated via behavior cloning with log loss:
 \begin{eqnarray}
\begin{split}
\mathcal{L}_{\text{bc}}=\mathbb{E}_{\tau\sim \mathcal{D}}\left[\frac{1}{T}\sum_{t=1}^T-\frac{1}{n}\sum_{i=1}^n \log \pi_\theta(u_t^i|o_{1:t}^i)\right],
\end{split}
\label{eq:bc}
\end{eqnarray}
where $\tau=\{o_t^i,u_t^i,r_t\}_{t=1:T}^{i=1:n}$ is the joint trajectory sampled from the dataset $\mathcal{D}$.In the experiment, we find the agent directly trained with behavior cloning tends to only choose the build-in action. This is because there is a serious class imbalance problem in the dataset, i.e., the number of build-in actions is much larger than other actions. To relieve the class imbalance problem, we introduce a weighting factor into the log loss. Similar to focal loss~\citep{lin2017focal}, we use an $\alpha$-balanced variant of the log loss:
\begin{eqnarray}
\begin{split}
\mathcal{L}_{\alpha-\text{balance}}=\mathbb{E}_{\tau\sim \mathcal{D}}\left[\frac{1}{T}\sum_{t=1}^T-\frac{1}{n}\sum_{i=1}^n \alpha(u^i)\log\pi_\theta(u_t^i|o_{1:t}^i)\right],
\end{split}
\label{eq:alpha_balance}
\end{eqnarray}
where $\alpha(u^i)$ is the weighting factor function, which is conditioned on the action label $u^i$. Besides, there is a designated player among the ten players who should not use the build-in action, so we add the extra loss to minimize the log probability of the build-in action of the designated player:
 \begin{eqnarray}
\begin{split}
\mathcal{L}_{\text{min\_build\_in}}=\mathbb{E}_{\tau\sim \mathcal{D}}\left[\frac{1}{T}\sum_{t=1}^T\sum_{i=1}^n -I(i,t)\log\left(1-\pi_\theta(u_\text{build\_in}^i|o_{1:t}^i)\right)\right],
\end{split}
\label{eq:min_buildin}
\end{eqnarray}
where $u_\text{build\_in}^i$ is represented for the build-in action and $I(i,t)$ is an indicator function, which is defined as:
 \begin{eqnarray}
\begin{split}
I(i,t)=\left\{
\begin{array}{rcl}
0 & & \text{if agent $i$ is a non-designated player at time step $t$}\\
1 & & \text{if agent $i$ is a designated player at time step $t$}.
\end{array} \right.
\end{split}
\label{eq:indicator}
\end{eqnarray}

Using past good experiences as the supervision has shown great success in many control tasks~\citep{oh2018self,vinyals2019grandmaster,zha2021simplifying}. Similar to the ranking buffer trick as proposed by ~\cite{zha2021simplifying}, we weight each trajectory in the dataset via their cumulative rewards. And trajectories with higher cumulative rewards receive higher weight and contribute more in the training loss:
 \begin{eqnarray}
\begin{split}
\mathcal{L}_{\text{buffer\_ranking}}=\mathbb{E}_{\tau\sim \mathcal{D}}\left[\rho(\tau)\frac{1}{T}\sum_{t=1}^T-\frac{1}{n}\sum_{i=1}^n \alpha(u^i)\log\pi_\theta(u_t^i|o_{1:t}^i)\right].
\end{split}
\label{eq:buffer_ranking}
\end{eqnarray}
where $\rho(\tau)$ is the trajectory weighting factor. And the practical setting of the trajectory weighting factor can be found in the Appendix.

To drive a policy improvement, we further adopt the Advantage-Weighted Regression (AWR)~\citep{peng2019advantage}. We first train a centralized value network $V_\phi^{\mathcal{D}}(s_{1:t})$ using the rewards from the dataset. And then compute the advantage with cumulative rewards (or reward return) $\mathcal{R}^\mathcal{D}$:
\begin{eqnarray}
\begin{split}
A(\mathbf{u}_t,s_{1:t}) = \mathcal{R}^\mathcal{D}-V_\phi^{\mathcal{D}}(s_{1:t}).
\end{split}
\label{eq:advantage}
\end{eqnarray}
Then the advantage weight is defined as:
\begin{eqnarray}
\begin{split}
w(\mathbf{u}_t,s_{1:t}) = \text{clip}\left(\exp\left(\frac{1}{\beta}A(\mathbf{u}_t,s_{1:t})\right),w_{\min},w_{\max}\right),
\end{split}
\label{eq:advantage_weight}
\end{eqnarray}
where $\beta>0$ is a fixed temperature parameter and $\text{clip}(\cdot)$ and we utilize the clipping function to limit the weight to the range of $w_{\min}$ to $w_{\max}$. The clipping function is helpful to relieve gradient explosion and gradient vanishing problems. The  advantage weight will be applied to the training loss:
\begin{eqnarray}
\begin{split}
\mathcal{L}_{\text{adv}}=\mathbb{E}_{\tau\sim \mathcal{D}}\left[\rho(\tau)\frac{1}{T}\sum_{t=1}^T-\frac{w(\mathbf{u}_t,s_{1:t})}{n}\sum_{i=1}^n \alpha(u^i)\log\pi_\theta(u_t^i|o_{1:t}^i)\right],
\end{split}
\label{eq:AWR_loss}
\end{eqnarray}

 In conclusion, the final training loss of our offline algorithm is the combination of the advantage-weighted loss and build-in action minimize loss:
\begin{eqnarray}
\begin{split}
\mathcal{L}_{\text{total}}= \mathcal{L}_{\text{adv}}+\eta\mathcal{L}_{\text{min\_build\_in}},
\end{split}
\label{eq:final_loss}
\end{eqnarray}
where $\eta$ is a loss balancing coefficient (i.e., a fixed hyper-parameter) and the loss is optimized with the Adam optimizer. In the next section, we will introduce how to train our offline RL algorithm in a practical distributed training framework.

\begin{figure*}[t]
\center
\includegraphics[width=1\linewidth]{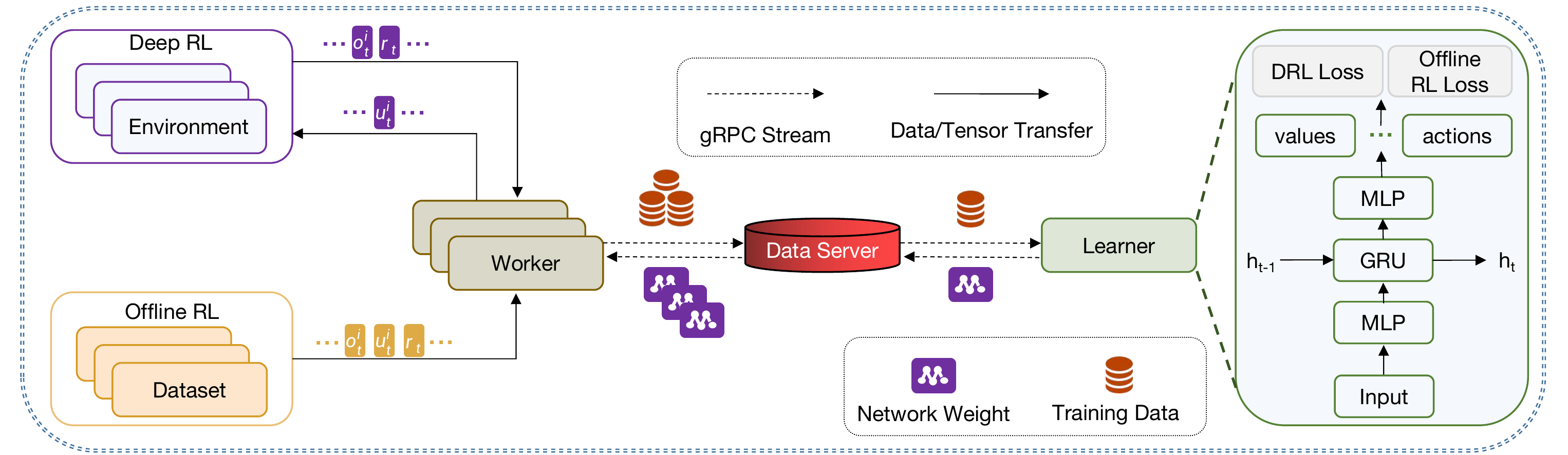}
\caption{The overall structure of the distributed training framework. The worker modules are used for producing training data. In offline reinforcement learning, the function of the work module is loading data from the disk and producing RNN hidden states. In deep reinforcement learning, the main function of the worker is interacting with the environment and producing actions and RNN hidden states. All the data collected by the worker modules will be sent to a data server via a gRPC stream. Then the learner modules will fetch data from the data server via a gRPC stream (the dotted arrow). After the training step, the weights will be sent to the worker modules through the data server.}
\label{fig:framework}
\end{figure*}

\subsection{Distributed Training Framework}

Training a powerful deep model needs a large-scale computation. In this work, we build a unified distributed training framework for both offline reinforcement learning and deep reinforcement learning. There are two main types of modules in our framework, i.e., the {\bf worker} modules and the {\bf learner} modules. The worker modules are used for producing necessary training data for the learner modules. In offline reinforcement learning, the main function of the work module is loading data from the disk and also producing RNN hidden states. In deep reinforcement learning, the main function of the worker module is interacting with environments (or simulators) to collect observations and rewards and also producing actions and RNN hidden states. In our distributed framework, a single worker module can load the data in parallel or interact with hundreds of environments. We can also launch multiple worker modules independently to further improve the throughput. All the data collected by the worker modules will be sent to a data server via a gRPC~\citep{grpc} stream. Then the learner modules will fetch data from the data server via a gRPC stream and start to train with pre-defined losses. In this work, we use the GRU unit to extract historical information. In the learner module, the update step of the GRU is set to 25, and the initialized GRU hidden states are produced by the worker modules. After the training step, the weights will be sent to the worker modules through the data server. Figure~\ref{fig:framework} shows the overall structure of our distributed training framework and Pytorch~\citep{paszke2019pytorch} is used for deep model building and parameter optimizing.

\section{Experiments}
In this section, we will evaluate our algorithm on the GRF full game and also show how the pre-trained model can accelerate multi-agent reinforcement learning in academic scenarios.

\subsection{Multi-Agent Google Research Football Full Game Evaluation}

In the multi-agent GRF full game, as shown in Figure~\ref{fig:screenshots}(a), we need to control all the players except the goalkeeper. To the best of our knowledge, TiKick is the first learning-based AI system that can take over the multi-agent GRF full game, while previous work could either control a single agent or experiment on toy academic scenarios. To evaluate our proposed offline reinforcement learning algorithm, we construct a series of incremental baselines from the vanilla multi-agent behavior cloning (MABC) algorithm.

{\bf Baselines:}

{\bf CQL}: CQL~\citep{kumar2020conservative} is an offline reinforcement learning algorithm that tries to learn conservative Q-values via adding penalties on the Q-functions.

{\bf MABC}: The multi-agent behavior cloning algorithm has been described in Section~\ref{sec:offline} and uses a naive supervised loss as shown in Equation~\ref{eq:bc}.

{\bf + $\alpha$-Balance} (abbreviated as +$\alpha$): Add an $\alpha$-balance weight to the MABC baseline to relieve the class imbalance problem  as shown in Equation~\ref{eq:alpha_balance}.

{\bf + Min Build-in}  (abbreviated as +MinBuild-in): Add a build-in action minimization loss to the previous baseline to force the designated player to choose the non-build-in action as shown in Equation~\ref{eq:min_buildin}.

{\bf + Buffer Ranking}  (abbreviated as +BR): Add the buffer ranking trick to take advantage of past good experiences as shown in Equation~\ref{eq:buffer_ranking}.

{\bf + Advantage Weight}  (abbreviated as +AW): Add an advantage-weighted loss to the previous baseline and obtain the final loss as shown in Equation~\ref{eq:final_loss}. And this method has been served as the final TiKick model.

\begin{table*}[t]
\begin{center}
        \begin{tabular}{|r|c|c|c|c|c|c|}
            \hline
            & CQL & MABC  & {\bf +}$\alpha$ & {\bf +}MinBuild-in & {\bf +}BR & {\bf +}AW\\
            \hline
            Win Rate & 0 & 0.834  & 0.890 & 0.926  & 0.930 & {\bf 0.944} \\
            Draw Rate & 0.258 & 0.112 & 0.088 & 0.064  & 0.058 & {\bf 0.036} \\
            Failure Rate & 0.742 & 0.054 & 0.022 & {\bf 0.010} & 0.012 & 0.02 \\
            Goal Diff & -1.496 & 2.140 & 2.732 & 3.056 & 2.816 & {\bf 3.096} \\
            TrueSkill & ${6.33\pm 3.16}$ & ${19.33\pm3.21}$  &  ${20.88\pm3.20}$  & ${22.18\pm3.23}$  & ${22.39\pm3.23}$  &  ${\bf 22.84}\pm 3.25$ \\
            \hline
        \end{tabular}
    \end{center}
    \caption{Evaluation results of different algorithms on the multi-agent GRF full game. Results show that our final algorithm (+AW) achieves the best performance with the highest rate of winning and the highest goal difference. The results of the TrueSkill evaluation are shown in the last row and mean scores with the standard deviation are reported. Our final TiKick model achieves the highest TrueSkill score.}
    \label{table:full_game_eval}
\end{table*}
\begin{figure}[t]
\center
\subfloat[]{\begin{centering}
\includegraphics[width=0.48\linewidth]{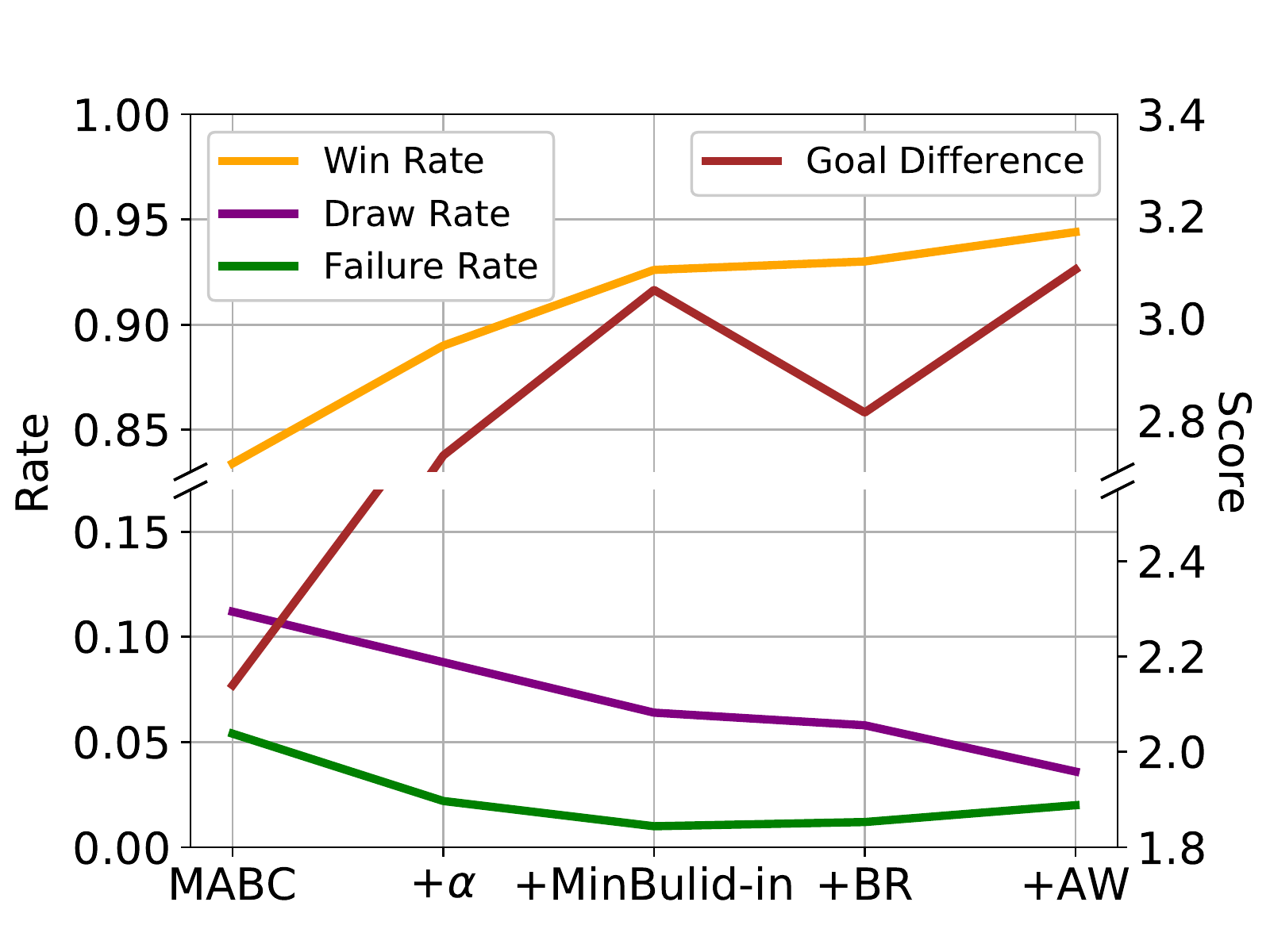}
\end{centering}
}\subfloat[]{\begin{centering}
\includegraphics[width=0.52\linewidth]{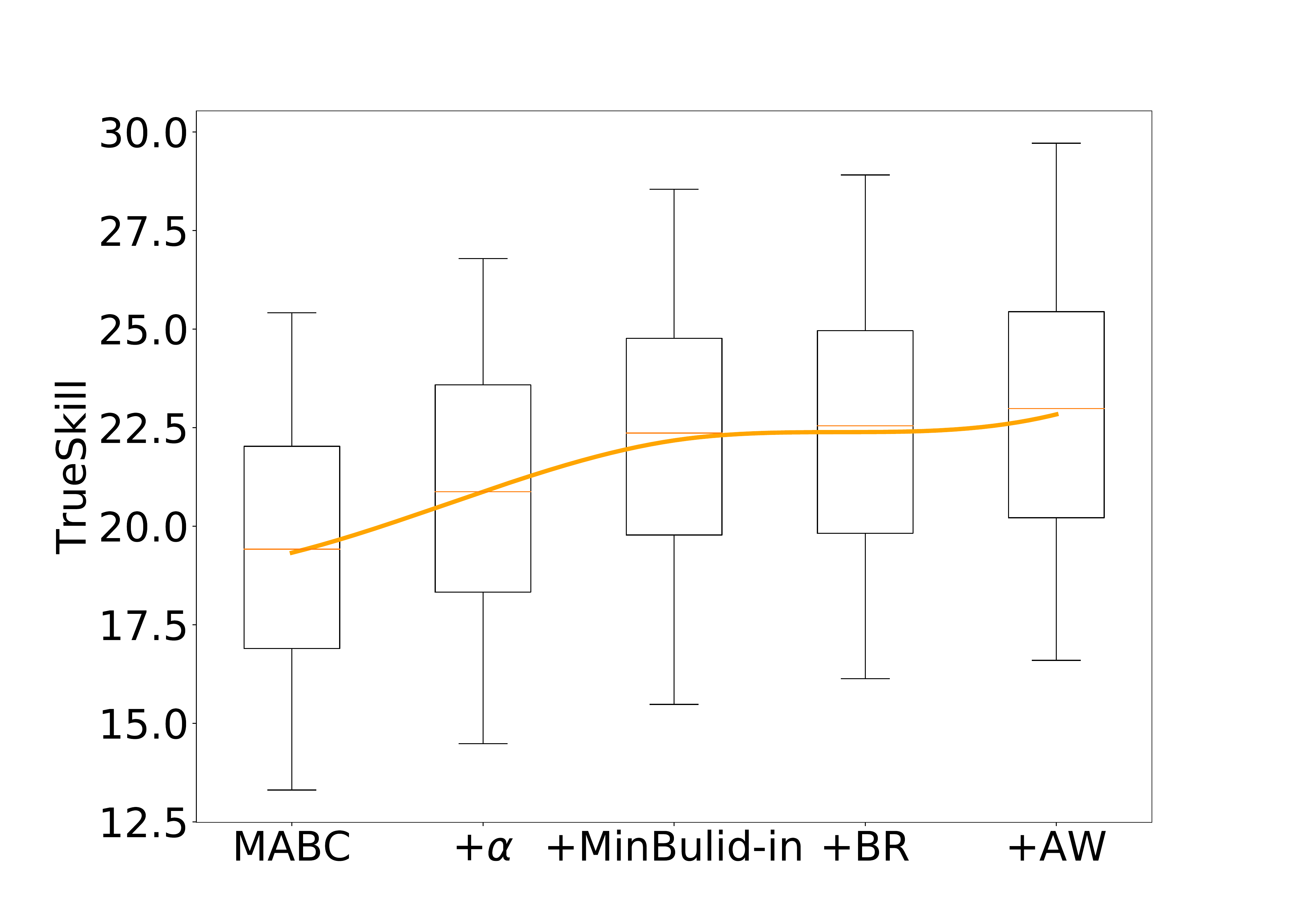}
\end{centering}
}
\caption{(a) The curves of the changes on the rate of winning, draw, and failure, and the goal differences of different algorithms. And the curve of the changes on the goal differences of different algorithms. Our final algorithm (+AW) achieves the best performance with the highest rate of winning and the highest goal difference. (b) TrueSkill evaluation for different algorithms. And our final TiKick model achieves the highest TrueSkill score.}
\label{fig:rate_and_elo}
\end{figure}

All the baselines share the same set of hyper-parameters and the same neural network backbone. Table~\ref{table:full_game_eval} shows the evaluation results of different algorithms on the multi-agent GRF full game. We report the rate of winning, draw, and failure, and also the goal differences after combating with the build-in AI for 500 rounds (with $3,000$ steps per round). Results show that CQL fails to defeat the build-in AI with a zero winning rate and our final algorithm (i.e., {\bf + Advantage Weight}) achieves the best performance with the highest rate of winning and the highest goal difference. Figure~\ref{fig:rate_and_elo}(a) shows the curves of the changes in the rate of winning, draw, and failure of different algorithms, and the curves of the changes on the goal differences of different algorithms. To further evaluate the performance of different algorithms, we rank all the algorithms with the TrueSkill rating system~\citep{herbrich2006trueskill}, and the results are shown in the last row of Table~\ref{table:full_game_eval} and Figure~\ref{fig:rate_and_elo}(b). And our final TiKick model achieves the highest TrueSkill score.

\begin{figure*}[t]
\includegraphics[width=1\linewidth]{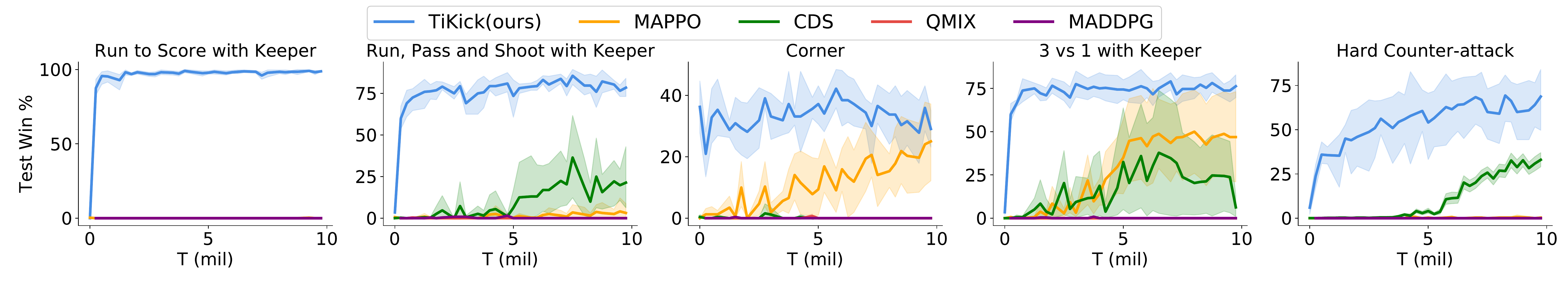}
\caption{Comparison of our method against baseline algorithms on GRF academic scenarios. Results show that our method, denoted as TiKick, achieves the best performance and the lowest sample complexity on all the scenarios.}
\label{fig:training_curves}
\end{figure*}

\begin{figure*}[t]
\center
\includegraphics[width=1\linewidth]{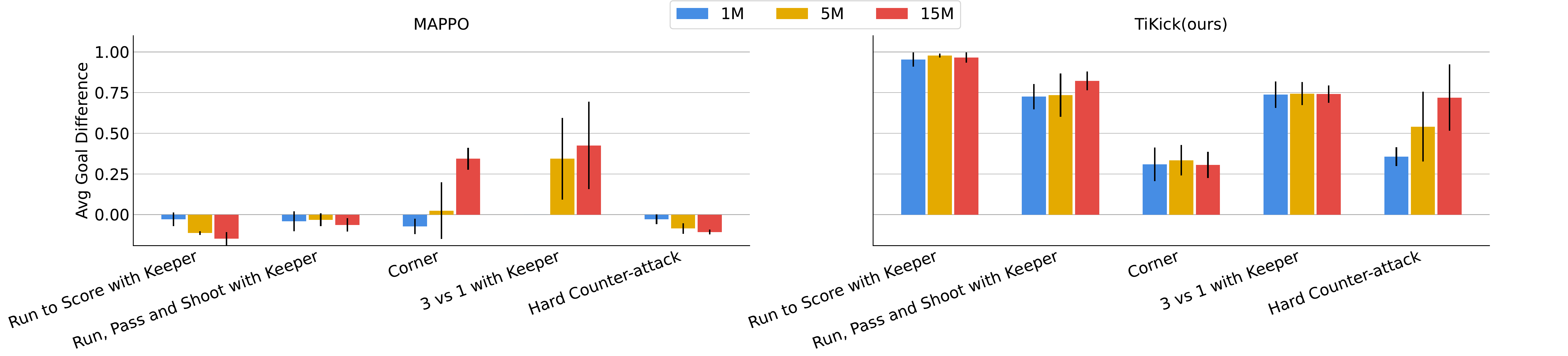}
\caption{The average goal difference on all the academic scenarios for MAPPO (left) and TiKick (right). Results show that our method achieves the highest scores on all the scenarios and our method can reach the highest scores within just 1M steps on four of five scenarios.}
\label{fig:histogram}
\end{figure*}

\subsection{Accelerate Multi-agent Reinforcement Learning Training}
Preivous work~\citep{silver2016mastering,vinyals2019grandmaster,ye2020supervised} shows that pre-trained supervised models help a lot for the follow-up deep reinforcement learning training. In this section, we will examine whether our pre-trained offline RL models can accelerate the multi-agent reinforcement learning training process. Thus, we consider five GRF academic scenarios originally proposed in \cite{googlefootball}, including:

\begin{itemize}
    \item[-] {\bf Run\_to\_Score\_with\_Keeper}
    \item[-] {\bf Run\_Pass\_and\_Shoot\_with\_Keeper}
    \item[-] {\bf Corner}
    \item[-] {\bf 3\_vs\_1\_with\_Keeper}
    \item[-] {\bf  Hard\_Counter-attack}
\end{itemize}

Figure~\ref{fig:screenshots}(b)-(f) show the screenshots of these five scenarios. In this experiment, we use QMIX~\citep{rashid2018qmix}, MADDPG~\citep{lowe2017multi}, CDS~\citep{li2021celebrating}  and MAPPO~\citep{yu2021surprising} as the baselines. 

{\bf Baselines:}

{\bf QMIX}: A value-based method that learns a monotonic approximation for the joint action-value function. QMIX factors the joint-action into a monotonic nonlinear combination of individual utilities of each agent. A mixer network with nonnegative weights is responsible for combining the agent’s utilities.

{\bf MADDPG}: A policy-based multi-agent algorithm, which adapts the single-agent DDPG algorithm to the multi-agent setting. 

{\bf CDS}: A value-based multi-agent algorithm with mutual information maximization between agents' identities and their trajectories. To be clear, CDS only combats with three enemies in the {\bf  Hard\_Counter-attack} scenario, while our method can combat with eleven enemies, which is much harder than the four-enemy version.

{\bf MAPPO}: A policy-based method that uses a centralized value function without individual utilities. MAPPO adapts the single-agent PPO algorithm to the multi-agent setting. Many useful tricks are used to stabilize the training, including Generalized Advantage Estimation (GAE)~\citep{schulman2015high}, PopArt~\citep{hessel2019multi}, observation normalization, value clipping, layer normalization, and ReLU activation with orthogonal initialization. 

Our method is built on the MAPPO algorithm and we load weights from our pre-trained TiKick offline RL model. Because there is a minor difference between the GRF full game and academic scenarios (i.e., academic scenarios have no build-in action), we do not load the weights of the last layer in the pre-trained model. In the experiment, we measure the winning rates and goal differences against the build-in AI over 5 seeds for all the algorithms and each evaluation data point is obtained from 64 test rounds.

We show the training performance comparison against baselines in Figure~\ref{fig:training_curves}. Results show that our method, denoted as TiKick, achieves the best performance and the lowest sample complexity on all the scenarios. Figure~\ref{fig:histogram} shows the average goal difference on all the academic scenarios for MAPPO and TiKick. Results show that our method achieves the highest scores on all the scenarios and our method can reach the highest score within just 1M steps on four of five scenarios. All the experiments show that the pre-trained offline RL model can accelerate the multi-agent RL training process.

\section{Conclusion}
In this paper, we have proposed a new multi-agent offline reinforcement learning method for the Google Research Football full game.  
Our proposed learning framework, denoted as TiKick, is implemented practically by using deep recurrent neural networks and carefully designed training losses. 
Experimental results show that our final algorithm achieves the best performance on the GRF full game. 
Furthermore, experiments on the GRF academic scenarios show that our pre-trained model can accelerate the multi-agent RL training process with better performance and lower sample complexity. 
In the future, we will try to apply our pre-trained model to the multi-agent league training to further improve the performance and robustness.

\bibliography{reference}
\bibliographystyle{nips}

\appendix

\section{Observation Design}
In this section, we will describe our observation design in detail. Our observation consists of the following parts:

\begin{table}[h]
\centering
\begin{tabular}{|l|}
\hline
absolute coordinates of players at our side\\
\hline
moving directions of players at our side\\
\hline
absolute coordinates of enemies\\
\hline
moving directions of enemies\\
\hline
absolute coordinate of current player\\
\hline
absolute coordinate of the ball\\
\hline
moving direction of the ball\\
\hline
one-hot indicator for the ball-owned team\\
\hline
one-hot indicator for the active player\\
\hline
one-hot indicator for the game mode\\
\hline
sticky actions\\
\hline
distance between the ball and current player\\
\hline
tired factors of players of our side\\
\hline
yellow card indicator of our side\\
\hline
red card indicator of our side\\
\hline
offside indicator of our side\\
\hline
offside indicator of enemies\\
\hline
distance between the enemies and current player\\
\hline
relative coordinates of players at our side\\
\hline
relative coordinates of players at our side\\
\hline
relative coordinates of players at our side\\
\hline
number of steps left till the end of the match\\
\hline
number of steps left till the end of the half-time\\
\hline
number goal differences\\
\hline
zero vector of 27 dimensions (used for future usage)\\
\hline
\end{tabular}
\end{table}
\newpage
\section{Hyper-parameters}
In this section, we will describe the hyper-parameters in detail. 
\begin{table}[h]
\centering
\begin{tabular}{cc}
\toprule
Hyper-parameters     & Value \\
\hline
policy network learning rate & $1e-4$\\
critic network learning rate & $1e-4$\\
number of environment parallel threads & 64 \\
number of MLP layers & 4\\
number of GRU layers & 1\\
training episode length & 1,500\\
test episode length & 3,000\\
data chunk length & 25\\
observation size & 268\\
action size & 20 for the GRF full game, 19 for academic scenarios\\
number of evaluation round & 500 for the GRF full game, 64 for academic scenarios\\
\hline
\end{tabular}
\end{table}

\section{Trajectory Weighting Factor Design}
In this section, we will describe how we design the trajectory weighting factor. Given a joint trajectory $\tau=\{o_t^i,u_t^i,r_t\}_{t=1:T}^{i=1:n}$ sampled from the dataset, we can calculate the cumulative reward as:
\begin{eqnarray}
\begin{split}
\mathcal{R}(\tau)=\sum_{t=1}^T r_t.
\end{split}
\label{eq:cumulative_reward}
\end{eqnarray}

Then we define a simple trajectory weighting factor as:
\begin{eqnarray}
\begin{split}
\rho(\tau) = \left\{
\begin{array}{rcl}
\rho_0 & & \text{if $\mathcal{R}(\tau)<\mathcal{R}_\mathrm{threshold}$}\\
\rho_1 & & \text{if $\mathcal{R}(\tau)\geq\mathcal{R}_\mathrm{threshold}$},
\end{array} \right.
\end{split}
\label{eq:weighting_factor}
\end{eqnarray}
where $\rho_0,\rho_1$ and $\mathcal{R}_\mathrm{threshold}$ are hyper-parameters. In this paper, we set $\rho_0=0,\rho_1=1$ and $\mathcal{R}_\mathrm{threshold}=3$.


\end{document}